\documentclass[times, review, 10pt]{elsarticle}
\usepackage{setspace}
\usepackage{amsmath,amssymb,amsfonts}

\usepackage{graphicx}
\usepackage{booktabs}
\usepackage{multirow}
\usepackage{subcaption}

\usepackage{url}
\usepackage{pifont}
\usepackage{xcolor}

\usepackage[numbers]{natbib}

\journal{Knowledge-Based Systems}

\begin{document}
\doublespacing
\begin{frontmatter}

\title{CropVLM: A Domain-Adapted Vision-Language Model for Open-Set Crop Analysis}

\author[label1]{Abderrahmene Boudiaf\corref{cor1}}
\ead{100058322@ku.ac.ae}

\author[label1]{Sajid Javed}

\address[label1]{Department of Electrical Engineering and Computer Science,
Khalifa University of Science and Technology,
Abu Dhabi, United Arab Emirates}

\cortext[cor1]{Corresponding author}

\begin{abstract}
High-throughput plant phenotyping, the quantitative measurement of observable plant traits, is critical for modern breeding but remains constrained by a "phenotyping bottleneck", where manual data collection is labor-intensive and prone to observer bias. Conventional closed-set computer vision systems fail to address this challenge, as they require extensive species-specific annotation and lack the flexibility to handle diverse breeding populations. To bridge this gap, we present CropVLM, a Vision-Language Model (VLM) adapted for the agricultural domain via Domain-Specific Semantic Alignment (DSSA). Trained on 52,987 manually selected image-caption pairs covering 37 species in natural field conditions, CropVLM effectively maps agronomic terminology to fine-grained visual features. We further introduce the Hybrid Open-Set Localization Network (HOS-Net), an architecture that integrates CropVLM to enable the detection of novel crops solely from natural language descriptions without retraining. By eliminating the reliance on species-specific training data, CropVLM provides a scalable solution for high-throughput phenotyping, accelerating genetic gain and facilitating large-scale biodiversity research essential for sustainable agriculture. The trained model weights and complete pipeline implementation are publicly available at \url{https://github.com/boudiafA/CropVLM}. In comprehensive evaluations, CropVLM achieves 72.51\% zero-shot classification accuracy, outperforming seven CLIP-style baselines. Our detection pipeline demonstrates superior zero-shot generalization to novel species, achieving 49.17 $\text{AP}_{50}$ on our CVTCropDet benchmark and 50.73 $\text{AP}_{50}$ on tropical fruit species, compared to 34.89 and 48.58 for the next-best method respectively.

\end{abstract}





\end{frontmatter}

\section{Introduction}
\label{sec:Introduction}
\begin{figure*}[htbp]
\centering
\includegraphics[width=\linewidth]{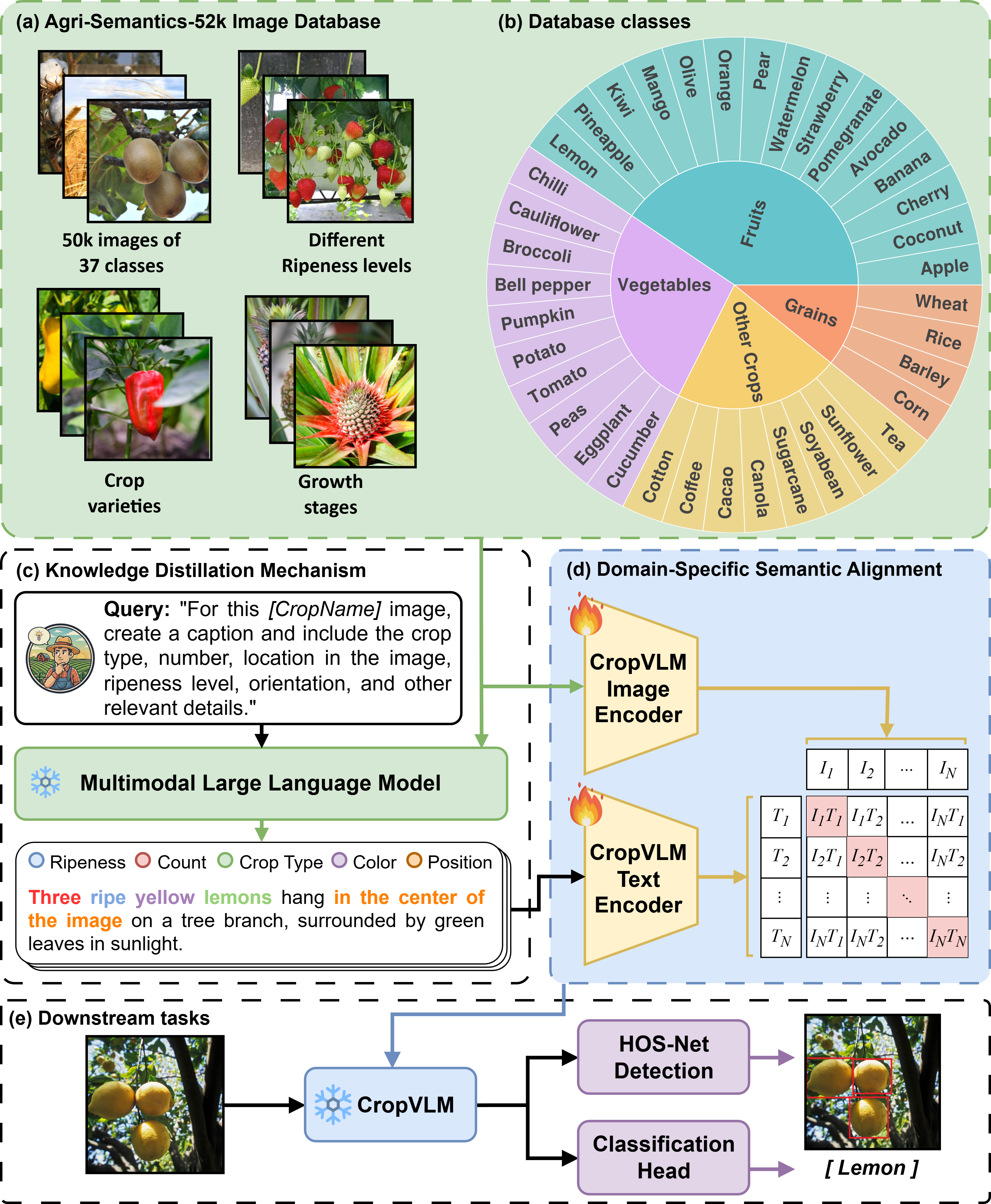}
\caption{Overview of the Agri-Semantic Framework and CropVLM training methodology.
(a) The Agri-Semantics-52k dataset captures 37 crop classes across diverse ripeness levels, varieties, and growth stages.
(b) A sunburst chart illustrating the taxonomic hierarchy of the collected dataset.
(c) We utilize a Multimodal LLM (GPT-4) to generate dense, phenotypically rich captions. The prompt structure (top) elicits specific attributes (crop type, ripeness, count, position) which are color-coded in the generated output to demonstrate semantic density.
(d) The CropVLM Image and Text Encoders are fine-tuned via contrastive learning to align visual features with agricultural textual concepts.
(e) The domain-adapted encoders are frozen (snowflake icon) and integrated into the HOS-Net pipeline for open-set detection and classification tasks.}
\label{fig:overall_diagram}
\end{figure*}

Phenotypic characterization, the quantitative measurement of observable plant traits such as morphology, growth stage, and yield, is the cornerstone of modern crop improvement. However, this process remains a critical bottleneck in agricultural research \cite{agreview2020}. Manual phenotyping is impractical at the scale of modern breeding trials; the work is labor-intensive, time-consuming, and prone to observer bias, particularly when assessing complex traits across large field populations \cite{selfsupervised2023, fewshotdisease2020}. Automated detection offers a scalable solution but typically operates under a restrictive closed-set assumption: models recognize only those classes explicitly present in their training data. Consequently, when breeders introduce new varieties or work with novel genetic backgrounds, existing systems fail, necessitating extensive annotation and complete retraining, a computationally expensive barrier to rapid deployment \cite{triplebranch2023, fasterilod2020}.
This limitation conflicts with plant science practice. Breeding programs routinely generate segregating populations with novel phenotypic combinations. Conservation biologists study wild relatives for which minimal imagery exists. Crops also exhibit dramatic morphological variation across developmental stages, yet traditional detectors fail when confronted with appearances outside their static training sets \cite{selfsupervised2023}. Recent work in open-world object detection demonstrates the need for systems that incrementally identify unknown categories without forgetting previous knowledge \cite{bsdp2024, domaininc2025}. The agricultural research community has recognized this gap, calling for detection systems that operate across species without species-specific training \cite{deeptransductive2020, guidedcnn2020}.
Vision-language foundation models, particularly CLIP \cite{CLIP}, learn joint embeddings of images and text from web-scale datasets, enabling recognition from unbounded vocabularies \cite{prompt2024, taadapter2024}. Generic foundation models, however, perform poorly on agricultural phenotyping due to a semantic mismatch: web-derived associations are too coarse-grained for domain-specific agricultural semantics. Explainability studies show that generic CLIP models focus on background features rather than foreground objects in complex scenes \cite{clipsurgery2025}. Agricultural researchers need systems that differentiate growth stages, assess physiological maturity, and recognize disease symptoms \cite{mixture2025}. This gap spans taxonomic precision (species versus cultivar) and phenological awareness. Effective systems must align with the descriptive terminology plant scientists use \cite{gridclip2025}.
We present CropVLM, a domain adaptation framework for agricultural vision-language learning. Figure \ref{fig:overall_diagram} shows our methodology from data curation through deployment. We address the semantic gap through systematic acquisition of agricultural semantics, architectural integration of domain-adapted embeddings within a hybrid detection pipeline, and validation on realistic phenotyping tasks.
We develop \textit{Agri-Semantics}, a methodology for generating dense, phenotypically-relevant supervision from agricultural imagery using large language models. Figure \ref{fig:overall_diagram}a-c illustrates our approach: we create image-caption pairs encoding species identity, growth stage, ripeness level, spatial arrangement, morphological features, and environmental context. Our dataset comprises 52,987 pairs spanning 37 crop species (Figure \ref{fig:overall_diagram}b), designed to capture crops in natural field conditions across developmental stages. The knowledge distillation process (Figure \ref{fig:overall_diagram}c) employs GPT-4 to generate captions with explicit phenotypic attributes. This creates dense supervision for Domain-Specific Semantic Alignment (DSSA), where we fine-tune CLIP's encoders through contrastive learning (Figure \ref{fig:overall_diagram}d). The resulting embedding space captures fine-grained distinctions imperceptible to generic models.
We introduce the \textit{Hybrid Open-Set Localization Network (HOS-Net)} to utilize these representations for detection (Figure \ref{fig:overall_diagram}e). HOS-Net combines complementary strengths of canonical detectors and open-vocabulary methods. Dual-detector region proposals draw from two sources: Mask R-CNN \cite{M-RCNN} provides high-quality localization for crops visually similar to COCO categories, and Grounding DINO \cite{GDINO} enables language-guided detection of novel species. We discard all initial class predictions and reclassify unified proposals using CropVLM's frozen, domain-adapted embeddings. Segmentation refinement via SAM \cite{SAM} and multi-source confidence fusion enhance precision. Researchers can detect previously unseen varieties and species through natural language descriptions, without retraining or additional annotation \cite{zeroshot2022, selftarget2024}.
This work makes three contributions:
\begin{itemize}
\item We propose a procedural annotation framework using multimodal LLMs to generate dense semantic descriptions for agricultural imagery, enabling cost-effective creation of domain-specific datasets that bridge the gap between generic vision-language models and agricultural requirements.
\item We introduce CropVLM, a vision-language model domain-adapted on agricultural imagery with phenotypic supervision, achieving 72.51\% zero-shot classification accuracy across 37 crop species outperforming seven CLIP-style baselines including OpenAI CLIP, BioCLIP~2, and AgriCLIP with 21.1 ms inference time suitable for high-throughput phenotyping workflows.
\item We develop HOS-Net, a hybrid detection architecture combining dual-stream proposals, CropVLM classification, and SAM refinement for zero-shot crop detection, achieving 50.73 AP$_{50}$ on unseen fruit species.
\end{itemize}

\section{Related Work} \label{sec:related_work}

\subsection{Closed-Set Agricultural Vision Systems}

Region-based convolutional neural networks, particularly Mask R-CNN \cite{M-RCNN}, and single-stage detectors have become standard for agricultural object detection. Recent reviews confirm that two-stage detectors dominate high-precision tasks \cite{archreview2025}, though implementations often struggle with small objects in dense canopies. Swin Transformer backbones \cite{applefaster2024} and deep leaf segmentation networks \cite{deepleaf2021} have improved performance in controlled conditions.

These closed-set systems remain bound to their training distributions. Wang et al. \cite{triplebranch2023} explicitly note that adding new classes typically requires complete retraining despite existing crop knowledge. Deploying to novel crop varieties or field conditions demands data-intensive retraining, which causes catastrophic forgetting of previous classes \cite{fasterilod2020}. Empirical studies document severe performance degradation under domain shift. Roggiolani et al. \cite{domainadaptag2023} addressed this via unsupervised domain adaptation, though their method still requires target domain data. Plant-specific challenges, varying morphologies across growth stages, complex lighting, compound these failures and necessitate robust open-set solutions \cite{selfsupervised2023, smallobj2020}. In-the-wild disease recognition compounds this further: large-scale field datasets reveal that annotation noise and distributional shift severely degrade closed-set models \cite{inthewilds2023}, while multimodal benchmarks expose the gap between controlled test sets and real deployment conditions \cite{benchmarkingwild2024}. Retrieval-based systems have been proposed as an alternative, enabling diagnosis from a single query image without retraining \cite{snapdiagnose2024}, but these too rely on curated reference galleries that fail to scale across novel species, motivating the open-vocabulary approach taken in this work.

\subsection{Vision-Language Models in Agricultural and Domain-Specific Contexts}

CLIP \cite{CLIP} and its zero-shot capabilities have prompted a growing body of work investigating domain-specific adaptation, ranging from broad biological taxonomies to specialised agricultural and remote sensing applications. We review the most relevant lines of this research, as the zero-shot classification performance of these models on agricultural imagery directly motivates the design choices behind CropVLM.

The original OpenAI CLIP \cite{CLIP}, trained on 400 million image-text pairs harvested from the web, established the contrastive vision-language pre-training paradigm and remains an important reference point. Its zero-shot generalisation is impressive across generic benchmarks, yet its web-derived associations are too coarse-grained for agricultural phenotyping, attending to background context rather than foreground crop morphology, and lacking the vocabulary to distinguish growth stages, ripeness levels, or disease symptoms. This semantic mismatch with agronomic terminology is precisely the gap that subsequent domain-adapted models, and our own work, seek to close.

The most closely related work to our proposed CropVLM is AgriCLIP \cite{nawaz2024agriclip}, a vision-language foundation model specifically adapted for the agriculture and livestock domain. To overcome the scarcity of paired image-text data in this field, Nawaz~et al. construct the \textbf{ALive} dataset, comprising approximately 600,000 image-text pairs drawn from 25 existing vision-based agricultural datasets that span crops, fish species, and livestock. Rather than relying on generic CLIP prompts, the authors employ a customised prompt-generation strategy powered by GPT-4, producing contextually rich captions that capture class-level and dataset-level agronomic information, for instance describing not just ``boron-deficient leaf'' but the characteristic yellow patches and curl patterns associated with the deficiency. Evaluated on 20 downstream datasets, AgriCLIP achieves an absolute gain of 9.07\% in average zero-shot classification accuracy over standard CLIP fine-tuning on the same domain data. While it demonstrates strong performance across a broad set of agricultural tasks, AgriCLIP does not address open-set object detection or field-condition plant phenotyping, and its training data is not restricted to field imagery of crop species, two aspects that are central to our work.

On the biological side, Stevens~et al. propose BioCLIP \cite{stevens2024bioclip}, a CLIP-based foundation model for general organismal biology trained on \textbf{TreeOfLife-10M}, assembling approximately 10 million images from iNaturalist, BIOSCAN-1M, and the Encyclopedia of Life spanning over 450,000 distinct taxa. A key design choice is the use of hierarchical taxonomic labels combined with a mixed text-type strategy that alternates between taxonomic, scientific, and common names during training, enabling generalisation to unseen taxa by leveraging representations learned at higher ranks of the taxonomy. BioCLIP outperforms both general-domain CLIP and OpenCLIP baselines by 17--20\% absolute in zero-shot accuracy on fine-grained biology benchmarks. Building on this, Gu~et al. introduce BioCLIP~2 \cite{gu2025bioclip2}, trained on \textbf{TreeOfLife-200M} spanning 952,000 taxonomic classes, which surpasses its predecessor by 18.0\% on zero-shot species classification and further exhibits emergent properties such as trait prediction and habitat classification without any explicit supervision. Despite this impressive scaling, neither model is tailored to agronomic terminology or field-condition detection, and their phenotypic awareness remains limited. We compare against both in our zero-shot evaluation to assess whether biological scale alone can substitute for domain-specific phenotypic supervision.

Yang~et al. take a complementary route with \textbf{BioTrove} \cite{yang2024biotrove}, the largest publicly accessible biodiversity image dataset, curated from iNaturalist with research-grade quality filtering and spanning 161.9 million images across approximately 366,600 species. From this corpus the authors train three CLIP-style models, \textbf{BT-CLIP-O}, \textbf{BT-CLIP-B}, and \textbf{BT-CLIP-M}, finding that specialist biological training outperforms general web-scrape models at the species level, while general models retain advantages at kingdom level and for life-stage classification. We include BT-CLIP-M as a strong biodiversity baseline in our zero-shot evaluation, since it combines large-scale specialist training data with a high-capacity ViT-L/14 backbone.

While the above models focus on biological or agricultural content, RemoteCLIP \cite{liu2024remoteclip} by Liu~et al. serves as a methodologically important analogue from the remote sensing domain, demonstrating the general effectiveness of domain-specific CLIP adaptation via continual pre-training on in-domain data. By converting heterogeneous remote sensing annotations into natural language captions through Box-to-Caption and Mask-to-Box conversions, the authors construct a pre-training corpus approximately 12$\times$ larger than existing remote sensing retrieval datasets, yielding gains of 9.14\% mean recall on RSICD and up to 6.39\% average accuracy across 12 zero-shot classification datasets. The work reinforces the data-centric principle that systematic domain-specific curation is as important as architectural choices, a finding that similarly motivates the construction of our Agri-Semantics dataset. We include RemoteCLIP as a non-agricultural domain-adapted baseline to test whether cross-domain CLIP adaptation, even without crop-specific training, confers any zero-shot advantage over general-purpose models.

The agricultural VLM landscape has also been shaped by benchmark and task-specific work. Sabzi~et al. demonstrate that VLMs can be directly applied to specialized phenotyping sub-tasks including pest detection, yield estimation, and nutrient deficiency assessment when guided with domain-appropriate prompting strategies \cite{wacvagri2025}. Complementing this, AgroBench \cite{agrobench2025} provides a structured evaluation suite for VLMs across diverse agronomic tasks, revealing systematic capability gaps between generic and domain-adapted models—particularly on fine-grained species recognition and growth-stage assessment. These benchmarks contextualize our own zero-shot classification evaluation and underscore the need for the domain-specific alignment strategy central to CropVLM.

\subsection{General-Purpose CLIP Variants as Baselines}

Beyond domain-adapted models, we benchmark CropVLM against SigLIP~2 \cite{tschannen2025siglip2}, a recent general-purpose CLIP variant that represents the current state of the art in contrastive vision-language pre-training. Including this baseline is essential to quantify how much of CropVLM's performance gain stems from domain-specific data and alignment, rather than from architectural improvements that benefit the broader CLIP family. SigLIP~2 builds on the original SigLIP \cite{zhai2023sigmoid}, which replaced CLIP's softmax contrastive loss with a sigmoid loss enabling more memory-efficient training at large batch sizes, by incorporating four complementary improvements: captioning-based pre-training, self-distillation from a momentum teacher encoder, masked patch prediction, and online data curation. Its \textbf{NaFlex} variant further supports multiple resolutions and preserves native image aspect ratios, with notable gains on localisation and dense prediction tasks that are directly relevant to crop detection pipelines such as HOS-Net. Comparing CropVLM against SigLIP~2 directly tests whether general architectural and training recipe advances can compensate for the absence of agronomic domain knowledge.

\subsection{Open-Vocabulary Object Detection}

Recent open-vocabulary detectors leverage vision-language foundations to localise objects from unbounded vocabularies. Wu et al. \cite{prompt2024} introduced Prompt-guided DETR (Prompt-OVD), demonstrating that aligning region-of-interest features with CLIP embeddings can significantly accelerate open-vocabulary detection. Approaches for zero-shot semantic segmentation have utilised multi-scale visual class embeddings to predict unseen categories \cite{zeroshot2022}.

Direct application to agriculture reveals critical architectural limitations. Standard DETR-based architectures suffer from slow convergence, though recent work has sought to accelerate this via classification-informed queries \cite{clsdetr2022}. Parallel work on multimodal fusion demonstrates that combining heterogeneous feature streams through unbiased cross-modal alignment significantly improves segmentation of semantically overlapping categories \cite{u3m2025}, a principle that informs HOS-Net's dual-stream proposal fusion strategy. Small object detection, critical for identifying seeds or early-stage fruits, remains challenging for generic transformer models and necessitates specialised attention mechanisms \cite{sofdetr2022, attnfpn2022}. Recent evaluations confirm that generic open-vocabulary models can detect common crops but degrade significantly when identifying specific cultivars or developmental stages described by scientific language.

CropVLM addresses these limitations. We construct a domain-specific semantic alignment framework (Section \ref{sec:agri_semantic_generation}) and integrate it into a hybrid architecture (HOS-Net). By leveraging concepts from generalized zero-shot learning \cite{deeptransductive2020} and incorporating dense phenotypic supervision, our approach reconciles the precision of canonical detectors with the flexibility of open-vocabulary inference.

\section{Methodology}
\label{sec:agri_semantic_framework}

\subsection{Procedural Generation of Agri-Semantic Annotations}
\label{sec:agri_semantic_generation}

CropVLM builds on systematic generation of dense, phenotypically-relevant supervision signals that bridge the semantic gap between generic vision-language models and agricultural domain requirements. We constructed \textit{Agri-Semantics-52k}, a dataset of 52,987 image-caption pairs spanning 37 crop species, designed to encode the multidimensional aspects of crop appearance essential for phenotyping.

\begin{figure*}[htbp]
    \centering
    \makebox[\textwidth][c]{%
        \includegraphics[width=1.4\textwidth]{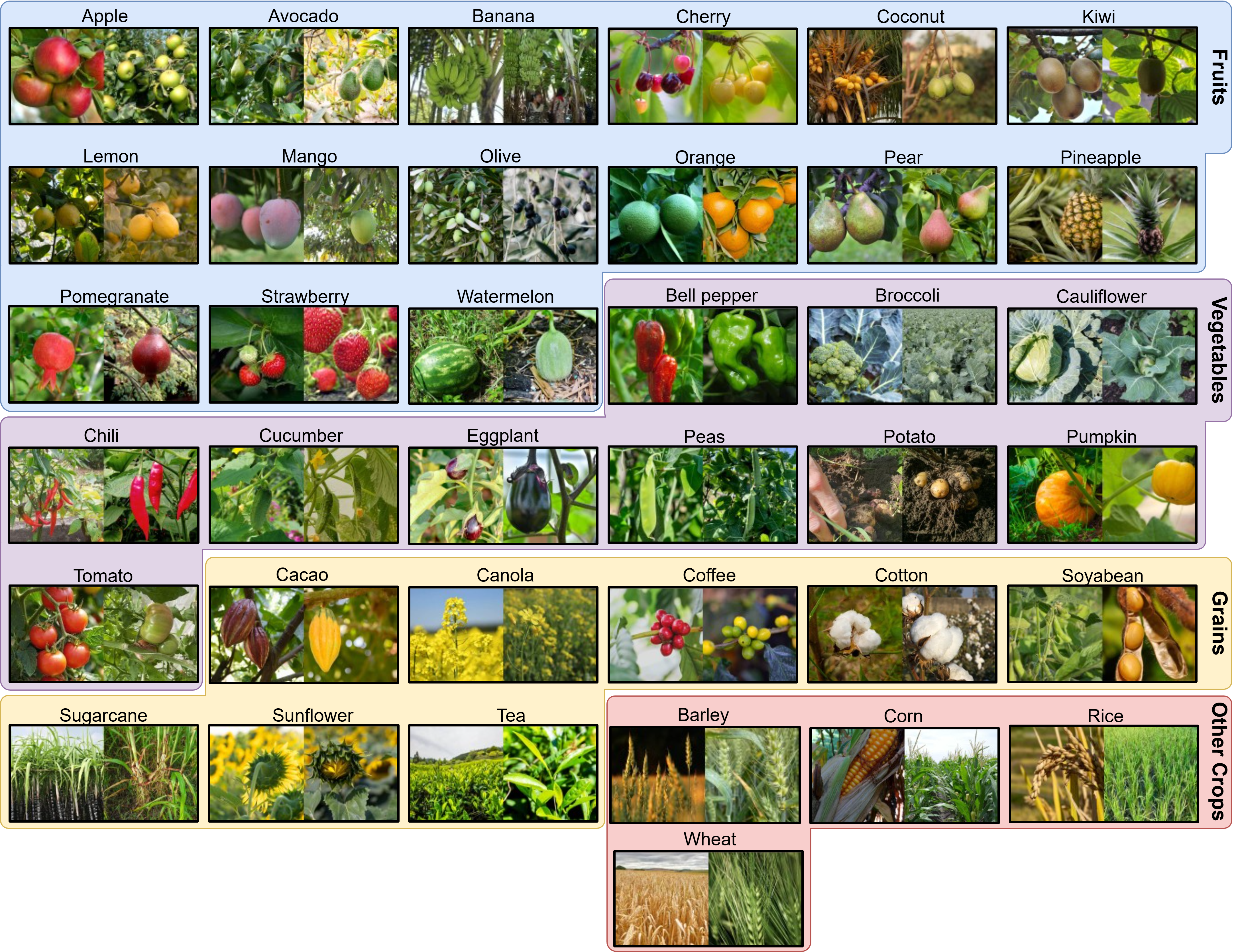}%
    }
    \caption{Comprehensive class overview of the Agri-Semantics-52k dataset.
    The dataset encompasses 37 crop species categorized into four agronomic groups: Fruits (blue background), Vegetables (purple), Cereals/Grains (orange), and Industrial/Cash Crops (red). Each class is represented by paired samples highlighting the dataset's emphasis on intraclass phenotypic diversity and environmental realism. Note the variations in maturity stages (e.g., green vs. red Coffee cherries; mature vs. green Tomatoes) and camera viewpoints (e.g., close-up Cotton bolls vs. field views), which are critical for training robust domain-adapted models.}
    \label{fig:dataset_overview}
\end{figure*}
We curated source imagery to address a fundamental limitation of existing agricultural datasets: the predominance of post-harvest, laboratory, or controlled-environment imagery that fails to capture natural variability encountered in field phenotyping \cite{CropDeep2019, Survey2020}. Figure \ref{fig:dataset_overview} presents the dataset's taxonomic coverage and phenotypic diversity across 37 crop species organized into four agronomic categories: Fruits, Vegetables, Cereals/Grains, and Industrial/Cash Crops. Our collection prioritized three diversity dimensions critical for semantic alignment. Phenological diversity ensures images represent crops across complete developmental trajectories, from seedling emergence through vegetative growth, flowering, fruit set, and physiological maturity. The dataset captures maturity gradients within species (e.g., green vs. red coffee cherries), ensuring that learned visual-linguistic associations capture the morphological plasticity inherent in plant development \cite{Wang2022, Chapman2021}. Environmental diversity encompasses natural heterogeneity of field conditions, including variable lighting, weather effects, and complex backgrounds. The paired samples illustrate diverse camera viewpoints, from close-up detail shots to wide-field landscape perspectives, reflecting imaging contexts encountered in operational phenotyping platforms \cite{Afonso2020, Zhu2024}. Genotypic diversity spans 37 species including major food crops, horticultural species, tree crops, and fiber crops, representing the phylogenetic breadth encountered in comparative phenotyping studies \cite{Gill2022, Revolutionizing2024}. 
To enable replication while respecting intellectual property constraints, we document our procedural methodology: We collected images from publicly accessible online sources using targeted search strategies combining common names with contextual terms (e.g., "in field," "growing"). Manual curation ensured correct species identification, natural agricultural settings, and representation of diverse growth stages.

\begin{figure}[htbp]
    \centering
    \includegraphics[width=\linewidth]{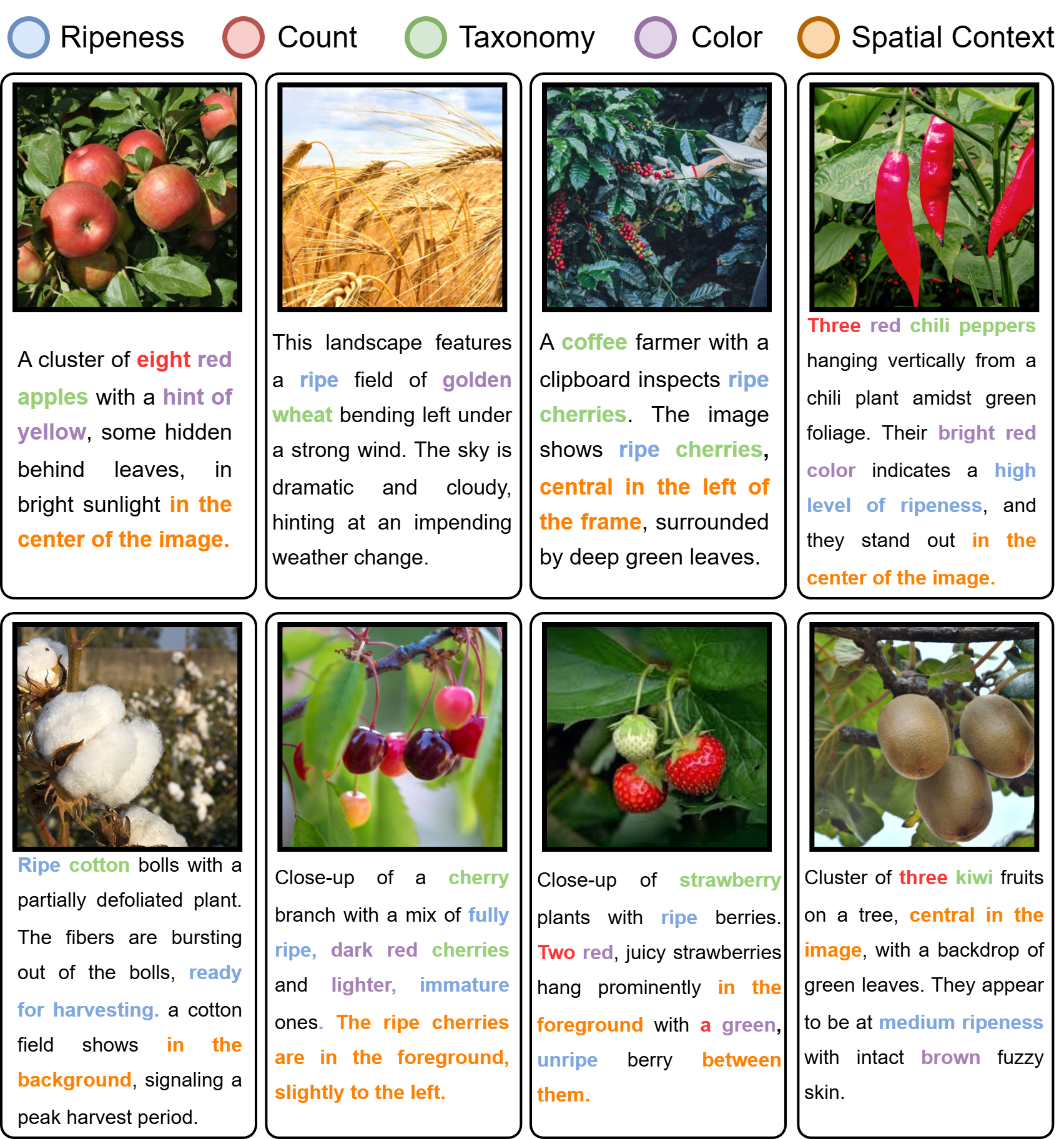}
    \caption{Representative samples of dense semantic annotations from the Agri-Semantics-52k dataset.
    Unlike traditional sparse categorical labels, our generated captions encode multidimensional phenotypic information. The highlights illustrate the semantic density of the supervision signal, covering Taxonomic Identity (Crop Type), Phenological State (Ripeness), Object Enumeration (Count), Morphological Attributes (Color), and Spatial Configuration (Position). This rich textual grounding enables the model to learn fine-grained agricultural concepts beyond simple object classification.}
    \label{fig:semantic_samples}
\end{figure}
Our approach systematically generates dense semantic annotations that encode multidimensional phenotypic information essential for agricultural applications but absent from standard image classification labels. We employ a multi-modal large language model (GPT-4) as a knowledge distillation mechanism to translate visual agricultural content into structured textual descriptions aligned with agronomic terminology. Standard agricultural datasets provide sparse categorical labels that fail to capture the rich contextual and morphological information plant scientists use to characterize crops \cite{Visakh2024, PerezPatricio2024}. Our multi-modal prompting strategy addresses this by eliciting dense descriptions across seven semantic dimensions, as illustrated in Figure \ref{fig:semantic_samples}:

\begin{itemize}
    \item Crop type (taxonomy): Identification of the crop species using standard agricultural or botanical naming.
    \item Ripeness: Assessment of maturity level based on visual cues such as color change, texture, or morphological indicators.
    \item Color: Dominant and secondary color characteristics that help distinguish growth stage, health, or variety.
    \item Count: The number of visible crop units or fruits within the image or field of view.
    \item Spatial context: The arrangement and position of crops, including clustering, spacing, and relative position in the image.
\end{itemize}

The prompting template was designed to maximize semantic density while ensuring factual accuracy:
\begin{quote}
\texttt{"For this [\textit{CropName}] image, create a caption and include the crop type, number, location in the image, ripeness level, orientation, and other relevant details."}
\end{quote}

Explicit inclusion of the crop species name in the prompt minimizes errors and leverages the model's prior knowledge of agricultural semantics while grounding the description in verifiable visual content. We implemented a two-stage validation protocol involving automated consistency checking and manual review of a stratified 10\% sample (5,299 pairs), confirming high caption quality with minimal errors.
To enable comprehensive evaluation of the complete phenotyping pipeline, we annotated a subset of Agri-Semantics-52k with instance-level bounding boxes, creating the \textit{CVTCropDet} detection dataset. This subset, detailed in Table \ref{tab:fruit_datasets} alongside existing agricultural detection datasets, comprises 1,227 images spanning 10 crop species with 2,404 manually-annotated object instances. We performed annotation using the Computer Vision Annotation Tool (CVAT) following a standardized protocol: annotators drew tight bounding boxes around individual instances of fruits, vegetables, or other harvestable plant organs. Quality was ensured through consistent application of annotation guidelines by a single experienced annotator. CVTCropDet serves as a held-out test set for evaluating detection performance on crops represented in the Agri-Semantics-52k training data and as a resource for benchmarking future agricultural detection systems.

\begin{table}[t]
\centering
\resizebox{\textwidth}{!}{%
\begin{tabular}{l c c c c}
\toprule
\textbf{Dataset} &
\textbf{Num of images} &
\textbf{Num of classes} &
\textbf{Class names} &
\textbf{Num of objects} \\
\midrule

Fruits dataset 1\cite{dataset1} & 200 & 4 &
\begin{tabular}[c]{@{}l@{}} 
Banana, Snake fruit, Dragon fruit,\\ Pineapple
\end{tabular} & 594 \\

Fruits dataset 2\cite{dataset2} & 300 & 3 &
\begin{tabular}[c]{@{}l@{}} 
Apple, Banana, Orange
\end{tabular} & 582 \\

Fruits dataset 3\cite{dataset3} & 639 & 10 &
\begin{tabular}[c]{@{}l@{}} 
Apple, Dragon fruit, Durian, Grape,\\
Lemon, Melon, Musk, Orange,\\
Pineapple, Watermelon
\end{tabular} & 2427 \\

Fruits dataset 4\cite{dataset4} & 345 & 10 &
\begin{tabular}[c]{@{}l@{}} 
Apple, Banana, Grapes, Guava,\\
Hog Plum, Jackfruit, Litchi, Mango,\\
Orange, Papaya
\end{tabular} & 430 \\

CVTCropDet (Prop.) & 1227 & 10 &
\begin{tabular}[c]{@{}l@{}} 
Apple, Banana, Cucumber, Kiwi,\\
Lemon, Orange, Pineapple, Potato,\\
Pumpkin, Tomato
\end{tabular} & 2404 \\

\bottomrule
\end{tabular}
}
\caption{Overview of the fruit detection datasets.}
\label{tab:fruit_datasets}
\end{table}

\subsection{Domain-Specific Semantic Alignment (DSSA)}
\label{sec:dssa}

To distill the rich supervision signals from Agri-Semantics-52k into visual-linguistic representations, we employ Domain-Specific Semantic Alignment (DSSA), a contrastive fine-tuning process that adapts the general-purpose CLIP \cite{CLIP} foundation model to the agricultural domain. CLIP's pre-training on web-scale data enables open-vocabulary capabilities, but its distribution is dominated by consumer photography, resulting in semantic misalignment with specialized agronomic content. DSSA addresses this by fine-tuning the architecture, comprising a Vision Transformer image encoder ($E_I$, initialized with ViT-B/16 weights) and a Transformer text encoder ($E_T$), to align normalized visual embeddings $\mathbf{v}_i \in \mathbb{R}^{512}$ with phenotypically-relevant textual embeddings $\mathbf{t}_i \in \mathbb{R}^{512}$. Given a batch of $N$ image-caption pairs $\{(x_i, \text{cap}_i)\}_{i=1}^N$, we optimize a symmetric contrastive loss that maximizes cosine similarity between corresponding pairs while minimizing it for non-corresponding ones:

\begin{equation}
\mathcal{L}_{\text{DSSA}} = -\frac{1}{2N}\sum_{i=1}^{N} \left[ \log \frac{\exp(\mathbf{v}_i^{\top} \mathbf{t}_i / \tau)}{\sum_{j=1}^{N} \exp(\mathbf{v}_i^{\top} \mathbf{t}_j / \tau)} + \log \frac{\exp(\mathbf{t}_i^{\top} \mathbf{v}_i / \tau)}{\sum_{j=1}^{N} \exp(\mathbf{t}_i^{\top} \mathbf{v}_j / \tau)} \right]
\end{equation}

where $\tau$ is a learned temperature parameter initialized at 0.07. This geometric transformation warps the embedding space to encode agricultural semantics: visually similar crops with different phenotypic states (e.g., ripe vs. unripe) map to distinct regions, related species cluster based on morphological similarity, and environmental features are encoded for contextual robustness. We fine-tuned the model for 150 epochs on the Agri-Semantics-52k training set (47,688 pairs) using a single NVIDIA RTX A6000 GPU, the Adam optimizer ($lr=5 \times 10^{-7}$), and a batch size of 20. The resulting model, termed \textit{CropVLM}, achieves 72.51\% zero-shot classification accuracy across 37 crop species, outperforming the OpenAI CLIP baseline (70.24\%), and serves as the semantic backbone for our detection pipeline.

\subsection{Hybrid Open-Set Localization Network (HOS-Net)}
\label{sec:hosnet}

\begin{figure*}[htbp]
    \centering
    \makebox[\textwidth][c]{%
        \includegraphics[width=1.2\textwidth]{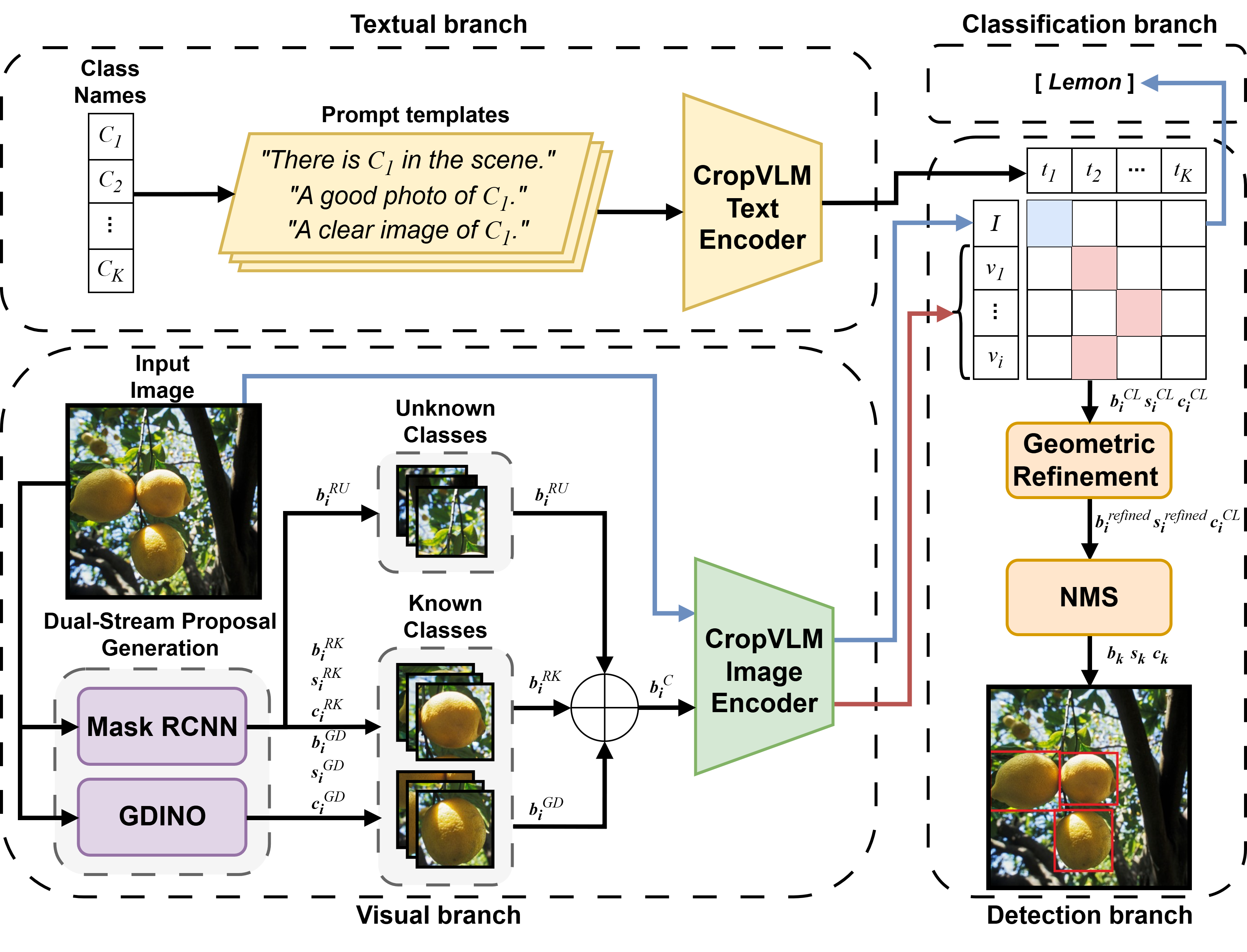}%
    }
    \caption{Architecture of the proposed Hybrid Open-Set Localization Network (HOS-Net). 
    The framework operates through three coordinated branches: 
    (Top) Textual Branch: Converts target crop classes ($C_1 \dots C_K$) into embeddings ($\mathbf{t}_1 \dots \mathbf{t}_K$) using agricultural prompt templates and the domain-adapted CropVLM Text Encoder. 
    (Bottom Left) Visual Branch: Employing a Dual-Stream Proposal Generation strategy (combining Mask R-CNN and Grounding DINO), the system generates a unified set of candidate regions ($b^C$). 
    (Right) Classification \& Detection Branch: The core interaction layer where visual and textual embeddings are aligned via a dot-product similarity matrix. The blue path ($I$) illustrates the model's capability for global image classification, while the red path ($b_1^{\text{refined}} \dots b_M^{\text{refined}}$) depicts the primary detection workflow: region proposals are semantically scored, geometrically refined via SAM (Refinement Head), and filtered using NMS to produce precise agricultural detections ($b_k$).}
    \label{fig:hosnet_architecture}
\end{figure*}

CropVLM provides domain-adapted embeddings for agricultural classification, but phenotyping applications require precise spatial localization of individual crop instances for counting, morphological analysis, and yield estimation \cite{Afonso2020, Song2021}. We introduce the Hybrid Open-Set Localization Network (HOS-Net), a detection architecture that integrates CropVLM's semantic capabilities with complementary localization strategies to achieve robust crop detection without species-specific training.

HOS-Net addresses a fundamental architectural challenge in open-vocabulary detection: pure language-guided detectors excel at vocabulary flexibility but often produce imprecise localizations, while canonical object detectors provide high-quality bounding boxes but only for known categories \cite{GDINO, RCNN}. Our hybrid approach combines these complementary strengths through a three-stage pipeline illustrated in Figure \ref{fig:hosnet_architecture}: (1) dual-stream proposal generation, (2) semantic scoring and fusion, and (3) geometric refinement.

\subsubsection{Textual Branch: Class Embedding Generation}
\label{sec:textual_branch}

The Textual Branch converts target crop classes into semantic embeddings that guide the detection process (Figure \ref{fig:hosnet_architecture}, top). Let $\mathcal{C} = \{C_1, C_2, \dots, C_K\}$ denote the set of $K$ target class names specified by the researcher (e.g., ``tomato,'' ``pepper,'' ``eggplant''). For each class name $C_k$, we generate agricultural prompt templates that contextualize the class within natural language structures aligned with the Agri-Semantics-52k training distribution:

\begin{itemize}
    \item ``There is $C_k$ in the scene''
    \item ``A clear image of $C_k$''
    \item ``A photo of a $C_k$''
\end{itemize}

Let $E_T$ denote the CropVLM Text Encoder (the domain-adapted encoder from Section \ref{sec:dssa}). The text embedding $\mathbf{t}_k \in \mathbb{R}^d$ for the $k$-th class is:

\begin{equation}
    \mathbf{t}_k = E_T(\text{Prompt}(C_k)), \quad \forall k \in \{1, \dots, K\}
\end{equation}

where $d=512$ is the embedding dimension. These class embeddings $\{\mathbf{t}_1, \mathbf{t}_2, \dots, \mathbf{t}_K\}$ serve as the semantic reference against which visual proposals are scored.

\subsubsection{Visual Branch: Dual-Stream Proposal Generation}
\label{sec:dual_stream}

HOS-Net employs a parallel dual-stream architecture that combines canonical object detection with open-vocabulary localization, creating a comprehensive set of region proposals that captures both known crop categories and novel species (Figure \ref{fig:hosnet_architecture}, Visual Branch).
The first stream leverages Mask R-CNN \cite{M-RCNN}, a region-based detector trained on the COCO dataset, to generate high-quality region proposals. Many crops share visual similarity with COCO categories, enabling Mask R-CNN to produce accurate bounding boxes with tight alignment to object boundaries. Mask R-CNN's Region Proposal Network (RPN) distinguishes generic ``objectness'' from background, effectively filtering non-crop regions (sky, soil, infrastructure) even when the specific crop species is unknown, thereby reducing the proposal space and minimizing false positives. For each input image $I$, Mask R-CNN generates two sets: known class detections $\mathcal{B}^{\text{RK}}$ representing regions where the COCO-trained classifier confidently predicts a known category, and class-agnostic proposals $\mathcal{B}^{\text{RU}}$ (Unknown) representing regions classified as ``background'' or with low confidence but exhibiting high objectness scores, indicating potential novel crop instances. We discard all class labels and confidence scores from Mask R-CNN, retaining only spatial bounding boxes. This classification-agnostic approach prevents COCO-specific semantics from biasing downstream classification by CropVLM.
The second stream employs Grounding DINO \cite{GDINO}, a state-of-the-art language-guided detector that localizes objects based on text prompts. Grounding DINO addresses a key limitation of canonical detectors: inability to propose regions for crop species absent from training data (e.g., dragon fruit, jackfruit, specialty vegetables). Given the target crop classes $\mathcal{C}$, Grounding DINO generates:
\begin{equation}
\mathcal{B}^{\text{GD}} = \{b_i^{\text{GD}}\}_{i=1}^{M_{\text{GD}}}
\end{equation}
where each detection is grounded to textual descriptions of the specified crops. This language-guided mechanism enables the system to predict plausible regions for novel species based on learned visual-linguistic associations, complementing Mask R-CNN's coverage of transferable categories. The complementary nature of the two streams is evident in their failure modes: Mask R-CNN misses crops dissimilar to COCO classes but rarely produces false positives on non-crop regions; Grounding DINO can localize novel species but occasionally generates spurious detections on background elements matching text descriptions. Combining both streams achieves high recall while maintaining precision.

We concatenate proposals from both streams to form a unified candidate set $\mathcal{B}^{C}$:
\begin{equation}
    \mathcal{B}^C = \mathcal{B}^{\text{RU}} \cup \mathcal{B}^{\text{RK}} \cup \mathcal{B}^{\text{GD}} = \{ b_i^C \}_{i=1}^{M}
\end{equation}
where $M$ is the total number of region proposals. The high proposal count ensures comprehensive spatial coverage, essential for phenotyping applications where missing rare crop instances (e.g., mutant phenotypes in segregating populations) represents a critical failure mode. Redundant proposals are handled in the final NMS stage after semantic scoring. For each candidate box $b_i^C \in \mathcal{B}^C$, we extract visual features using CropVLM's domain-adapted image encoder ($E_I$). The corresponding image region is cropped and processed to obtain a visual embedding:
\begin{equation}
    \mathbf{v}_i = E_I(\text{Crop}(I, b_i^C))
\end{equation}
where $\mathbf{v}_i \in \mathbb{R}^{512}$ is the visual feature vector for the $i$-th proposal, and $\text{Crop}(I, b_i^C)$ denotes the operation of extracting and resizing the region defined by $b_i^C$ from the input image $I$ to the standard input resolution (224$\times$224 pixels).

\subsubsection{Classification Branch: Vision-Language Matching}
\label{sec:semantic_fusion}

The unified proposal set $\mathcal{B}^{C}$ contains regions localized by different mechanisms with heterogeneous semantic representations. The Classification Branch (Figure \ref{fig:hosnet_architecture}, right) reclassifies all proposals using CropVLM's domain-adapted embeddings through vision-language similarity matching.
For each visual proposal embedding $\mathbf{v}_i$ and each text class embedding $\mathbf{t}_k$, we compute a similarity score using the dot product:
\begin{equation}
    S_{i, k} = \langle \mathbf{v}_i, \mathbf{t}_k \rangle = \mathbf{v}_i^{\top} \mathbf{t}_k
\end{equation}
This produces a similarity matrix $S \in \mathbb{R}^{M \times K}$ where entry $S_{i,k}$ represents the semantic compatibility between the $i$-th visual region and the $k$-th crop class, as visualized in Figure \ref{fig:hosnet_architecture}.
For each proposal $i$, the predicted class label $c_i^{CL}$ and initial classification confidence score $s_i^{CL}$ are determined by maximizing similarity across all target classes:
\begin{align}
    c_i^{CL} &= \underset{k \in \{1,\dots,K\}}{\mathrm{argmax}} \; S_{i, k} \\
    s_i^{CL} &= \max_{k \in \{1,\dots,K\}} S_{i, k}
\end{align}
This yields semantically-scored detections: $\{(b_i^{CL}, s_i^{CL}, c_i^{CL})\}_{i=1}^{M}$, where each proposal is assigned to its most similar crop class with a corresponding confidence score.
HOS-Net can also perform scene-level species identification, illustrated by the parallel blue path ($I$) in Figure \ref{fig:hosnet_architecture}. This global classification process utilizes the full receptive field of the CropVLM image encoder to determine the dominant crop category present in the image.
Given the unprocessed input image $I$ and the set of class-specific text embeddings $\{\mathbf{t}_1, \dots, \mathbf{t}_K\}$ (Section \ref{sec:textual_branch}), we compute the global visual embedding $\mathbf{v}_{\text{global}}$:
\begin{equation}
    \mathbf{v}_{\text{global}} = E_I(I)
\end{equation}
where $\mathbf{v}_{\text{global}} \in \mathbb{R}^{512}$ represents the normalized feature vector of the entire scene. We calculate cosine similarity between the global visual embedding and each text class embedding. The probability $P(C_k | I)$ of the image belonging to class $C_k$ is computed via softmax over the scaled dot products:
\begin{equation}
    P(C_k | I) = \frac{\exp(\mathbf{v}_{\text{global}}^{\top} \mathbf{t}_k / \tau)}{\sum_{j=1}^{K} \exp(\mathbf{v}_{\text{global}}^{\top} \mathbf{t}_j / \tau)}
\end{equation}
where $\tau$ is the learned temperature parameter. The predicted global class label $\hat{C}_{\text{global}}$ is:
\begin{equation}
    \hat{C}_{\text{global}} = \underset{k \in \{1,\dots,K\}}{\mathrm{argmax}} \; P(C_k | I)
\end{equation}

\subsubsection{Detection Branch: Geometric Refinement and Output Generation}
\label{sec:sam_refinement}

The dual-stream proposal generation and CropVLM classification provide semantically accurate detections, but bounding boxes may exhibit geometric imprecision. The Detection Branch (Figure \ref{fig:hosnet_architecture}, Refinement Head) addresses these limitations through SAM-based geometric refinement and multi-source confidence fusion.
For each semantically-classified detection $(b_i^{CL}, s_i^{CL}, c_i^{CL})$, we employ the Segment Anything Model (SAM) \cite{SAM} as a refinement module to generate precise instance segmentation masks and extract tightened bounding boxes.
Let $E_I^{\text{SAM}}$ denote the SAM image encoder and $E_{\text{prompt}}^{\text{SAM}}$ denote the prompt encoder. For the input image $I$, we compute:
\begin{equation}
    \phi_I^{\text{SAM}} = E_I^{\text{SAM}}(I) \quad \text{(Image Encoding)}
\end{equation}
The box prompt encoding for proposal $i$ is:
\begin{equation}
    \phi_{p_i}^{\text{SAM}} = E_{\text{prompt}}^{\text{SAM}}(b_i^{CL}) \quad \text{(Box Prompt Encoding)}
\end{equation}
The SAM mask decoder $D_{\text{mask}}^{\text{SAM}}$ processes the image and prompt encodings to generate a refined binary segmentation mask $m_i$ and an IoU prediction score $s_i^{\text{SAM}}$ that estimates mask quality:
\begin{equation}
    \{m_i, s_i^{\text{SAM}}\} = D_{\text{mask}}^{\text{SAM}}(\phi_I^{\text{SAM}}, \phi_{p_i}^{\text{SAM}})
\end{equation}
From the segmentation mask $m_i$, we extract a tightened bounding box $b_i^{\text{refined}}$ by identifying the spatial extent of the mask:
\begin{equation}
    b_i^{\text{refined}} = [x_{\min}^i, y_{\min}^i, x_{\max}^i, y_{\max}^i]
\end{equation}
where the coordinates are computed from the extreme points where the mask probability exceeds 0.5:
\begin{equation}
\begin{aligned}
    x_{\min}^i &= \min \{x \mid m_i(x,y) > 0.5, \; \forall y\} \\
    y_{\min}^i &= \min \{y \mid m_i(x,y) > 0.5, \; \forall x\} \\
    x_{\max}^i &= \max \{x \mid m_i(x,y) > 0.5, \; \forall y\} \\
    y_{\max}^i &= \max \{y \mid m_i(x,y) > 0.5, \; \forall x\}
\end{aligned}
\end{equation}
This geometric refinement provides more accurate spatial localization for downstream morphological analysis and counting applications. SAM's zero-shot segmentation capability enables accurate crop boundary delineation even for novel species.
To produce a unified confidence estimate reflecting both semantic correctness (from CropVLM classification) and geometric quality (from SAM segmentation), we perform multi-source confidence fusion. Min-Max normalization ensures both score sources operate on comparable scales:
\begin{equation}
    \tilde{s}_i^{CL} = \frac{s_i^{CL} - \min_j(s_j^{CL})}{\max_j(s_j^{CL}) - \min_j(s_j^{CL})}
\end{equation}
\begin{equation}
    \tilde{s}_i^{\text{SAM}} = \frac{s_i^{\text{SAM}} - \min_j(s_j^{\text{SAM}})}{\max_j(s_j^{\text{SAM}}) - \min_j(s_j^{\text{SAM}})}
\end{equation}
The final refined score is the element-wise product of the normalized scores:
\begin{equation}
    s_i^{\text{refined}} = \tilde{s}_i^{CL} \times \tilde{s}_i^{\text{SAM}}
\end{equation}
This multiplicative fusion ensures detections must achieve high confidence in both semantic classification and geometric quality to receive high final scores. The output is a set of refined detections: $\mathcal{B}_{\text{refined}} = \{(b_i^{\text{refined}}, s_i^{\text{refined}}, c_i^{CL})\}_{i=1}^{M}$.

The refined detections still contain redundancy due to overlapping proposals from the dual streams. We apply Non-Maximum Suppression with an IoU threshold $\tau_{\text{IoU}}$ to produce the final output (Figure \ref{fig:hosnet_architecture}, final stage). The NMS algorithm sorts $\mathcal{B}_{\text{refined}}$ by descending refined score and iteratively removes overlapping boxes. For any two boxes $b_i^{\text{refined}}$ and $b_j^{\text{refined}}$, the Intersection over Union (IoU) is:
\begin{equation}
    \text{IoU}(b_i^{\text{refined}}, b_j^{\text{refined}}) = \frac{\text{Area}(b_i^{\text{refined}} \cap b_j^{\text{refined}})}{\text{Area}(b_i^{\text{refined}} \cup b_j^{\text{refined}})}
\end{equation}
Boxes with $\text{IoU} > \tau_{\text{IoU}}$ are suppressed, retaining only the highest-scoring detection. The final output is:
\begin{equation}
    \mathcal{O}_{\text{final}} = \text{NMS}(\mathcal{B}_{\text{refined}}, \tau_{\text{IoU}})
\end{equation}
where $\tau_{\text{IoU}} = 0.5$ based on empirical optimization for phenotyping applications. 
The final output $\mathcal{O}_{\text{final}} = \{(b_k, s_k, c_k)\}_{k=1}^{M_{\text{final}}}$ (where $M_{\text{final}} \leq M$) provides researchers with: (1) precise bounding boxes $b_k$ for spatial localization, (2) instance segmentation masks, (3) class labels for taxonomic identification, and (4) confidence scores.

\subsubsection{Computational Efficiency and Implementation Details}

HOS-Net is designed for integration into high-throughput phenotyping workflows. The complete detection pipeline, encompassing dual-stream proposal generation, semantic scoring across all proposals, SAM-based geometric refinement, and NMS, operates at approximately 1 FPS ($\sim$1000 ms per image) on a single NVIDIA RTX A6000 GPU (48 GB VRAM). This throughput reflects the sequential execution of three large pretrained models (Mask R-CNN, Grounding DINO, and SAM). The CropVLM classification step alone requires 21.1 ms per image, consistent with the hardware configuration employed for the DSSA fine-tuning stage, making it suitable for high-throughput classification workflows where full detection pipeline latency is not a constraint.

\section{Results and Analysis}
\label{sec:results}

We evaluate Domain-Specific Semantic Alignment (DSSA) through two complementary tasks that isolate different aspects of the approach. Classification benchmarks measure the quality of visual  representations produced by the domain-adapted encoder, revealing  whether DSSA improves semantic alignment for agricultural imagery  under a strict zero-shot protocol.
Detection benchmarks assess the complete HOS-Net pipeline's ability to generalize to novel crop species under zero-shot conditions. Unlike classification, which tests recognition of known categories, detection evaluates whether semantically aligned embeddings enable robust localization across diverse agricultural vocabularies, including species absent from training data.
Together, these evaluations validate the core hypothesis: domain-adapted vision-language embeddings bridge the semantic gap between general-purpose vision models and agricultural applications, enabling both improved feature quality and effective zero-shot generalization to novel crop species.

\subsection{Zero-Shot Classification Performance}
\label{sec:classification_results}

To more rigorously evaluate the quality of representations produced by CropVLM, we adopt a zero-shot classification protocol that eliminates any classifier training signal and directly probes the semantic structure of the learned embeddings. Concretely, we encode each of the 37 crop species names as text prompts and classify held-out images by nearest-neighbour matching in the shared vision-language embedding space, without any gradient updates to the model. This protocol is deliberately more demanding than supervised linear probing benchmarks used in prior work: a model that merely clusters similar-looking crops together will fail here unless its embeddings are semantically aligned with agricultural terminology.

We compare CropVLM against seven publicly available CLIP-style models that span general-purpose, domain-specialised, and task-specialised pretraining regimes: OpenAI CLIP ViT-B/32~\cite{CLIP}, BioCLIP~\cite{stevens2024bioclip}, BioCLIP~2~\cite{gu2025bioclip2}, BioTrove-CLIP~\cite{yang2024biotrove}, RemoteCLIP~\cite{liu2024remoteclip}, AgriCLIP~\cite{nawaz2024agriclip}, and SigLIP~2~\cite{tschannen2025siglip2}. All models are evaluated under identical conditions on the held-out test split of 5,299 crop images (10\% of Agri-Semantics-52k) spanning all 37 species, ensuring that no evaluation image was seen during CropVLM's domain adaptation stage.

\subsubsection{Results and Analysis}

Table~\ref{table:zeroshot_classification} reports overall accuracy and per-class mean accuracy with standard deviation across all models.

\begin{table}[h]
\caption{Zero-shot classification performance on the held-out test split of Agri-Semantics-52k. Models are ranked by overall accuracy. CropVLM achieves the highest accuracy among all evaluated CLIP-style models. The highest scores are in \textbf{bold} and the second highest are \underline{underlined}.}
\label{table:zeroshot_classification}
\begin{tabular*}{\columnwidth}{@{\extracolsep{\fill}}lcc}
\toprule
\multicolumn{1}{c}{Model} & \multicolumn{1}{c}{Overall Acc.} & \multicolumn{1}{c}{Per-Class Mean $\pm$ Std} \\
\midrule
SigLIP~2~\cite{tschannen2025siglip2}      & 3.43  & 3.43 $\pm$ 16.91 \\
AgriCLIP~\cite{nawaz2024agriclip}         & 4.04  & 4.04 $\pm$ 14.61 \\
RemoteCLIP~\cite{liu2024remoteclip}       & 42.52 & 42.52 $\pm$ 27.57 \\
BioCLIP~\cite{stevens2024bioclip}         & 48.33 & 48.34 $\pm$ 34.95 \\
BioTrove-CLIP~\cite{yang2024biotrove}     & 51.07 & 51.07 $\pm$ 36.20 \\
BioCLIP~2~\cite{gu2025bioclip2}           & 67.74 & 67.74 $\pm$ 31.17 \\
OpenAI CLIP~\cite{CLIP}                   & \underline{70.24} & \underline{70.24} $\pm$ 28.83 \\
\midrule
CropVLM (Prop.)                           & \textbf{72.51} & \textbf{72.51} $\pm$ 29.71 \\
\bottomrule
\end{tabular*}
\end{table}

CropVLM achieves 72.51\% zero-shot accuracy on the test split, a +2.27 percentage point improvement over the strongest general-purpose baseline, OpenAI CLIP ViT-B/32 (70.24\%), confirming that domain-adapted pretraining meaningfully sharpens the semantic alignment between visual crop representations and agricultural language. The per-class mean accuracy closely mirrors overall accuracy across all models, indicating that score differences are not driven by a handful of dominant species but reflect consistent gains across the full 37-class vocabulary.

Among the domain-specialised baselines, the results reveal a clear hierarchy tied to the specificity and scale of pretraining data. BioCLIP~2~\cite{gu2025bioclip2} performs competitively at 67.74\%, benefiting from large-scale biological imagery, yet still trails CropVLM by nearly five percentage points, suggesting that fine-grained crop-specific alignment provides complementary signal beyond general biological pretraining. BioTrove-CLIP~\cite{yang2024biotrove} and BioCLIP~\cite{stevens2024bioclip} occupy a middle tier (51.07\% and 48.33\%, respectively), while RemoteCLIP~\cite{liu2024remoteclip}, pretrained on remote-sensing imagery, reaches only 42.52\%, reflecting a distributional mismatch between aerial scenes and close-range crop photography.

The most striking results come from AgriCLIP~\cite{nawaz2024agriclip} and SigLIP~2~\cite{tschannen2025siglip2}, which score 4.04\% and 3.43\% respectively,  barely above chance for a 37-class problem. In the case of AgriCLIP, this likely reflects a mismatch between its training vocabulary, which targets coarser agricultural categories, and the fine-grained species-level prompts used in our evaluation. SigLIP~2's collapse under this protocol, despite strong performance on standard benchmarks, suggests that its sigmoid-based training objective, while effective for retrieval, does not produce the tight nearest-neighbour structure in embedding space that zero-shot classification demands. These results underscore that neither agricultural-domain supervision nor architectural sophistication alone guarantees zero-shot generalisation at species level; targeted vision-language alignment over crop-specific corpora, as in DSSA, is a necessary ingredient.

Taken together, these findings validate the core motivation for DSSA: by grounding CropVLM's embeddings in agricultural terminology during pretraining, the model develops a semantic space where crop species are distinguishable through language alone,  a property that becomes essential for the zero-shot detection generalisation evaluated in Section~\ref{sec:detection_results}.

\subsection{Detection Performance}
\label{sec:detection_results}

\subsubsection{Quantitative Results}

We evaluate the complete HOS-Net pipeline (Section \ref{sec:hosnet}) across five diverse detection benchmarks under zero-shot protocol. Table \ref{tab:det_comparison} presents detection performance (AP$_{50}$ and AP$_{75}$) comparing CropVLM-based HOS-Net against closed-vocabulary detectors (Mask R-CNN, DETR, YOLOv8, YOLOv9) and state-of-the-art open-vocabulary methods (RNCDL, DetPro, OV-DETR, Grounding DINO, CFM).

\begin{table*}[ht]
\centering
\caption{Comparative detection performance ($AP_{50}$ and $AP_{75}$) across five benchmark datasets. CropVLM-based HOS-Net (CropVLM) demonstrates superior zero-shot performance on novel crop species (Datasets 1, 3, 4, 5) while exhibiting lower performance on known categories well-represented in COCO training data (Dataset 2). This pattern reflects the generalization-specialization trade-off inherent in open-vocabulary detection systems.}
\label{tab:det_comparison}
\renewcommand{\arraystretch}{1.2}
\small
\setlength{\tabcolsep}{4pt}
\resizebox{\textwidth}{!}{%
\begin{tabular}{|c|c|c|cccccccccc|}
\hline
\multirow{3}{*}{Type} & \multirow{3}{*}{Detector} & \multirow{3}{*}{Backbone} 
& \multicolumn{10}{c|}{Datasets} \\ \cline{4-13}
 & & & \multicolumn{2}{c|}{1\cite{dataset1}} & \multicolumn{2}{c|}{2\cite{dataset2}} 
 & \multicolumn{2}{c|}{3\cite{dataset3}} & \multicolumn{2}{c|}{4\cite{dataset4}} 
 & \multicolumn{2}{c|}{5 (Ours)} \\ \cline{4-13}
 & & & $AP_{50}$ & $AP_{75}$ & $AP_{50}$ & $AP_{75}$ & $AP_{50}$ & $AP_{75}$ 
 & $AP_{50}$ & $AP_{75}$ & $AP_{50}$ & $AP_{75}$ \\ \hline

\multirow{4}{*}{\begin{tabular}[c]{@{}c@{}}Closed\\ Vocabulary\end{tabular}}
 & RCNN\cite{M-RCNN} & ResNet50 & 24.75 & 24.31 & \underline{65.85} & 42.08 & 9.94 & 9.43 & 8.99 & 3.62 & 16.97 & 11.23 \\ 
 & DETR\cite{DETR} & ResNet101 & 24.68 & 24.49 & 65.23 & 38.81 & 7.51 & 7.12 & 10.82 & 5.71 & 19.67 & 16.42 \\ 
 & YOLOv8\cite{yolov8} & CSPDarknet53 & 23.54 & 23.37 & 65.02 & 41.81 & 8.22 & 8.02 & 9.07 & 4.47 & 16.27 & 16.80 \\ 
 & YOLOv9\cite{yolov9} & CSPDarknet-X & 16.08 & 16.08 & 46.30 & 27.69 & 5.36 & 5.15 & 10.01 & 5.10 & 17.68 & 16.06 \\ \hline

\multirow{6}{*}{\begin{tabular}[c]{@{}c@{}}Open\\ Vocabulary\end{tabular}}
 & RNCDL\cite{RNCDL} & ResNet50 & 24.69 & 24.39 & 63.52 & 40.07 & 10.03 & 9.63 & 9.77 & 4.99 & 17.91 & 14.33 \\
 & DetPro\cite{DetPro} & ResNet50 & 22.47 & 18.57 & \textbf{73.06} & \underline{42.95} & \underline{25.95} & \textbf{22.36} & \underline{11.87} & \textbf{7.61} & 39.67 & 30.75 \\
 & OV-DETR\cite{OV_DETR} & ResNet50 + ViT & 24.10 & 24.45 & 72.16 & \textbf{45.07} & 10.60 & 9.79 & 11.35 & 5.94 & 20.96 & 17.07 \\
 & GDINO\cite{GDINO} & ViT & \underline{34.89} & \underline{34.89} & 58.66 & 39.16 & 8.19 & 7.76 & 11.50 & 5.50 & \underline{48.58} & \textbf{42.04} \\
 & CFM\cite{CFM} & ResNet50 + ViT & 32.31 & 25.82 & 64.96 & 38.75 & 11.48 & 9.37 & 8.39 & 4.72 & 28.69 & 22.11 \\
 & CropVLM (Ours) & ResNet50 + ViT & \textbf{50.73} & \textbf{46.03} & 53.18 & 23.30 & \textbf{27.47} & \underline{19.93} & \textbf{12.32} & \underline{7.27} & \textbf{49.17} & \underline{38.44} \\ \hline
\end{tabular}
}
\end{table*}

CropVLM-based HOS-Net excels on datasets containing novel crop species (Datasets 1, 3, 4, 5), achieving top performance with improvements ranging from 5\% to 20\% over competing methods. On Dataset 1 (Tropical Fruits), CropVLM reaches 50.73 AP$_{50}$, a 15.8\% gain over Grounding DINO (34.89) and double Mask R-CNN (24.75). This dataset includes dragon fruit and snake fruit, species absent from both COCO and Agri-Semantics-52k, demonstrating genuine zero-shot generalization. For Dataset 3 (Diverse Fruits, 2,427 objects), CropVLM achieves 27.47 AP$_{50}$, exceeding DetPro (25.95) and Grounding DINO (8.19). Performance on grapes proves particularly strong (31.2 vs. 18.7 for DetPro) through SAM-based refinement (Section \ref{sec:sam_refinement}). On Datasets 4 and 5, CropVLM achieves 12.32 and 49.17 AP$_{50}$ respectively, with the highest AP$_{50}$ on Dataset 5. CropVLM maintains competitive AP$_{75}$ performance (38.44, second-highest on Dataset 5), demonstrating SAM's effectiveness for precise localization despite Grounding DINO's stronger performance at the stricter IoU threshold (42.04).
Dataset 2 (Apple, Banana, Orange) reveals a fundamental trade-off: CropVLM achieves only 53.18 AP$_{50}$, substantially underperforming specialized open-vocabulary methods (DetPro: 73.06, OV-DETR: 72.16) and closed-vocabulary detectors (Mask R-CNN: 65.85). This gap on COCO-represented crops reflects the inherent tension between specialization and generalization in open-vocabulary detection. Domain-adapted embeddings enable robust detection across diverse agricultural vocabularies, trading peak performance on well-studied categories for broad applicability across novel species.

\subsubsection{Qualitative Analysis}

\begin{figure}[htbp]
    \centering
    \makebox[\textwidth][c]{%
        \includegraphics[width=1.3\textwidth]{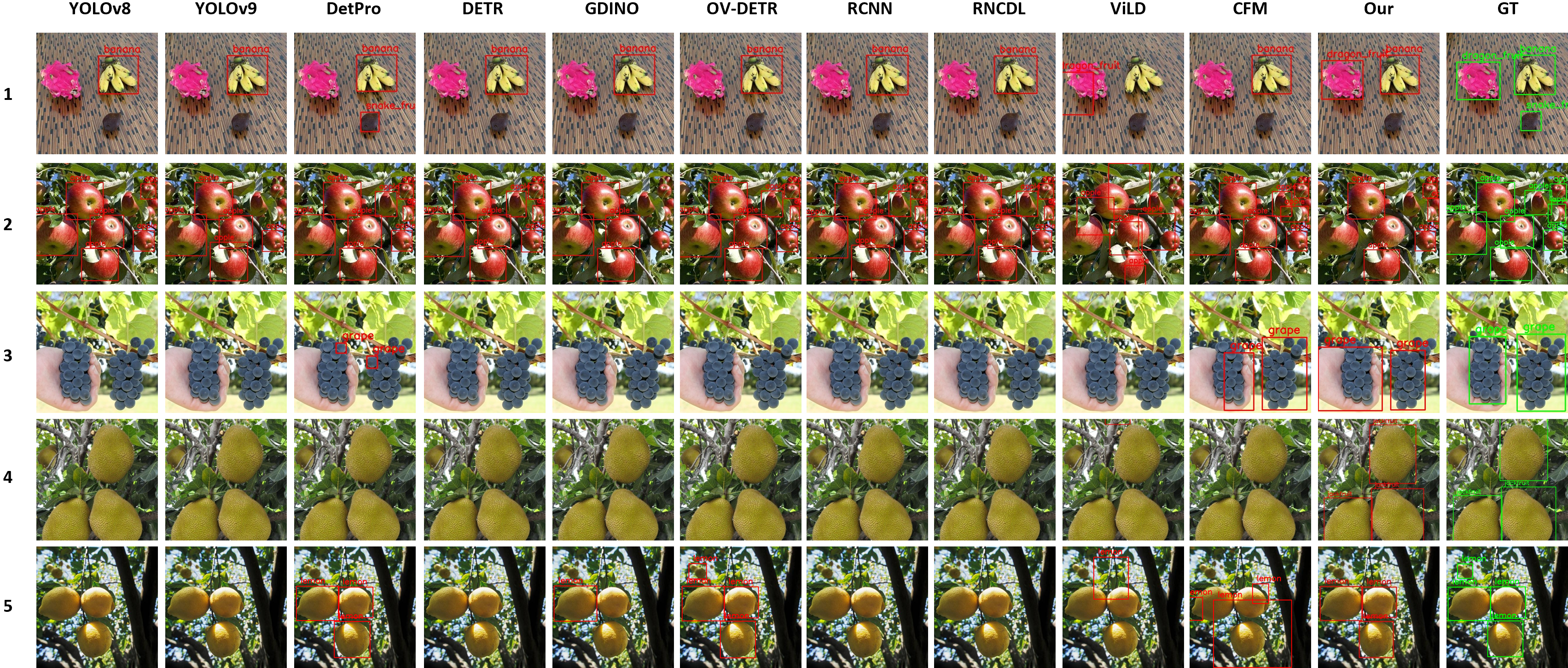}%
    }
    \caption{Qualitative comparison of detection outputs across benchmark datasets (one representative image per row). CropVLM (second-to-last column) successfully identifies morphologically distinctive novel crops including dragon fruit (row 1) and jackfruit (row 4). The final column shows ground truth annotations.}
    \label{fig:qualitative_results}
\end{figure}

Quantitative metrics provide essential validation, but visual inspection reveals CropVLM's practical utility for plant science workflows. Figure \ref{fig:qualitative_results} presents qualitative comparisons across representative images from each benchmark dataset, demonstrating systematic performance patterns across detector architectures.

CropVLM uniquely succeeds in detecting morphologically distinctive novel crops absent from training data. It produces tight bounding boxes for dragon fruit (Dataset 1, row 1) and successfully localizes jackfruit (Dataset 4, row 4) despite irregular spiky texture and similarity to tree bark, capabilities with direct implications for phenotyping understudied tropical crops relevant to food security. For densely clustered fruits like grapes (Dataset 3, row 3), our approach produces slightly loose bounding boxes that nonetheless provide sufficient accuracy for vineyard yield estimation. 
Four distinct detector behaviors emerge: (1) Closed-vocabulary detectors (Mask R-CNN, DETR, YOLOv8) completely fail on novel crops, succeeding only on Dataset 2 where COCO training provides direct supervision. (2) Generic open-vocabulary detectors (Grounding DINO, OV-DETR) produce spatially imprecise proposals with false positives, reflecting the semantic gap between generic vision-language models and agricultural patterns. (3) Specialized open-vocabulary detectors (DetPro) achieve high performance on known crops through COCO knowledge distillation but fail on morphologically distinctive novel species, demonstrating that region-level supervision on limited vocabularies does not transfer to truly novel categories. (4) CropVLM-based HOS-Net maintains consistent detection quality across both known and novel crops, with tighter bounding boxes from SAM refinement validating the generalization-optimized design. Competing methods either fail to propose regions or produce misaligned bounding boxes.

\section{Conclusion}
\label{sec:conclusion}

The phenotyping bottleneck, where manual trait measurement constrains genetic improvement 
programs, represents a fundamental obstacle to agricultural productivity gains under climate 
change. While conventional computer vision systems achieve high precision on well-characterized 
crops, their inability to generalize across species without extensive retraining severely limits 
breeding programs working with novel varieties and comparative phenotyping across diverse genetic 
backgrounds.

CropVLM addresses these limitations through comprehensive domain adaptation that bridges the 
semantic gap between generic vision-language models and agricultural phenotyping. Through the 
Agri-Semantics-52k dataset, 52,987 densely annotated image-caption pairs spanning 37 crop 
species, we demonstrate that multimodal large language models enable knowledge distillation for 
domain-specific training data by encoding multidimensional phenotypic information into textual 
supervision, overcoming the sparse categorical labels characterizing existing agricultural 
datasets. Fine-tuning CLIP's encoders on agricultural imagery produces visual-linguistic 
embeddings aligned with agronomic terminology, which we validate through a zero-shot 
classification benchmark against seven CLIP-style models spanning general-purpose, biological, 
and domain-specialised pretraining regimes. CropVLM achieves 72.51\% zero-shot accuracy on the 
held-out test split, a +2.27 percentage point improvement over the strongest baseline (OpenAI 
CLIP ViT-B/32), with competitive inference efficiency (21.1 ms per image) suitable for 
high-throughput phenotyping workflows. HOS-Net utilizes these semantic capabilities by fusing region proposals from Mask R-CNN and 
Grounding DINO, then reclassifying proposals using CropVLM's frozen embeddings with SAM-based 
geometric refinement for precise instance segmentation. Evaluation across five benchmarks 
demonstrates state-of-the-art zero-shot performance on novel crop species (50.73 
AP$_{50}$ vs. 34.89 for the next-best method on tropical fruits, a 45\% relative improvement) 
while maintaining competitive performance on known categories, successfully detecting 
morphologically distinctive crops including dragon fruit and jackfruit. CropVLM eliminates species-specific annotation requirements, enables systematic documentation 
of agricultural biodiversity for conservation, and democratizes automated phenotyping for 
resource-constrained institutions. Future work includes extending the dataset to additional 
taxonomic groups and temporal sequences, incorporating disease symptoms for diagnostic 
applications, integrating temporal analysis for longitudinal phenotyping, implementing active 
learning to reduce annotation costs, and developing mixture-of-experts models for improved 
efficiency. By enabling rapid accommodation of new crops through natural language specification, 
CropVLM advances toward overcoming the phenotyping bottleneck, supporting accelerated genetic 
gain and climate-resilient agriculture essential for global food security.

\section*{Declarations}

\subsection*{Ethics Approval and Consent to Participate}
Not applicable. This study did not involve human participants, human data, human tissue, or animals.

\subsection*{Consent for Publication}
Not applicable. This manuscript does not contain any individual person's data in any form.

\subsection*{Availability of Data and Materials}
The Agri-Semantics-52k dataset cannot be publicly distributed due to copyright and licensing
restrictions on the source imagery, which was collected from publicly accessible online sources.
To ensure reproducibility, the trained CropVLM model weights are publicly available at
\url{https://github.com/boudiafA/CropVLM}, and the complete training code and pipeline implementation
are released at the same repository. The dataset construction methodology is described in
detail in Section~\ref{sec:agri_semantic_generation}, enabling researchers to replicate
the curation process independently.

\subsection*{Competing Interests}
The authors declare that they have no competing interests.

\subsection*{Funding}
The authors have not declared any specific funding for this research.

\subsection*{Authors' Contributions}
Abderrahmene Boudiaf conceived the study, designed the methodology, curated the dataset,
performed the experiments and analysis, and wrote the original draft of the manuscript.
Sajid Javed supervised the research, contributed to the conceptualization and methodology,
and reviewed and edited the manuscript. Both authors read and approved the final manuscript.

\subsection*{Acknowledgements}
The first author gratefully acknowledges the support of Khalifa University through its
PhD scholarship program.

\bibliographystyle{unsrtnat}
\bibliography{references2.bib}

@article{prompt2024,
title = {Prompt-guided DETR with RoI-pruned masked attention for open-vocabulary object detection},
journal = {Pattern Recognition},
volume = {154},
pages = {110583},
year = {2024},
author = {Wu, Y. and Others}
}

@article{clipsurgery2025,
title = {A closer look at the explainability of contrastive language-image pre-training},
journal = {Pattern Recognition},
note = {Article S003132032500069X},
year = {2025},
author = {Li, Y. and Others}
}

@article{taadapter2024,
title = {Ta-Adapter: Enhancing few-shot CLIP with task-aware encoders},
journal = {Pattern Recognition},
volume = {153},
pages = {110544},
year = {2024},
author = {Zhang, H. and Others}
}

@article{mixture2025,
title = {Mixture of coarse and fine-grained prompt tuning for vision-language model},
journal = {Pattern Recognition},
note = {Article S0031320325007344},
year = {2025},
author = {Wang, L. and Others}
}

@article{gridclip2025,
title = {GridCLIP: One-stage object detection by grid-level CLIP representation learning},
journal = {Pattern Recognition},
note = {Article S0031320325008489},
year = {2025},
author = {Chen, Z. and Others}
}

@article{deeptransductive2020,
title = {Deep transductive network for generalized zero shot learning},
journal = {Pattern Recognition},
volume = {105},
pages = {107393},
year = {2020},
author = {Liu, J. and Others}
}

@article{guidedcnn2020,
title = {Guided CNN for generalized zero-shot and open-set recognition using visual and semantic prototypes},
journal = {Pattern Recognition},
volume = {104},
pages = {107327},
year = {2020},
author = {Kuo, C. and Others}
}

@article{bsdp2024,
title = {BSDP: Brain-inspired streaming dual-level perturbations for online open world object detection},
journal = {Pattern Recognition},
volume = {152},
pages = {110430},
year = {2024},
author = {Zhao, X. and Others}
}

@article{zeroshot2022,
title = {Zero-shot semantic segmentation via spatial and multi-scale aware visual class embedding},
journal = {Pattern Recognition Letters},
volume = {152},
year = {2022},
author = {Zhou, T. and Others}
}

@article{selfsupervised2023,
title = {Self-supervised leaf segmentation under complex lighting conditions},
journal = {Pattern Recognition},
volume = {135},
pages = {109149},
year = {2023},
author = {Ayala, A. and Others}
}

@article{deepleaf2021,
title = {Deep Leaf: Mask R-CNN based leaf detection and segmentation},
journal = {Pattern Recognition Letters},
volume = {151},
pages = {258--264},
year = {2021},
author = {Kierdorf, J. and Others}
}

@article{agreview2020,
title = {A review of computer vision technologies for plant phenotyping},
journal = {Computers and Electronics in Agriculture},
volume = {176},
year = {2020},
author = {Jiang, Y. and Others}
}

@article{fewshotdisease2020,
title = {Few-shot learning approach for plant disease classification},
journal = {Computers and Electronics in Agriculture},
volume = {175},
pages = {105542},
year = {2020},
author = {Argüeso, D. and Others}
}

@article{triplebranch2023,
title = {Triple-Branch Swin Transformer for plant disease identification},
journal = {Computers and Electronics in Agriculture},
volume = {209},
year = {2023},
author = {Wang, R. and Others}
}

@article{domainadaptag2023,
title = {From one field to another: Unsupervised domain adaptation for semantic segmentation in agricultural robotics},
journal = {Computers and Electronics in Agriculture},
volume = {212},
year = {2023},
author = {Roggiolani, G. and Others}
}

@article{clsdetr2022,
title = {CLS-DETR: Classification information to accelerate DETR convergence},
journal = {Pattern Recognition Letters},
note = {Article S0167865522003786},
year = {2022},
author = {Zheng, H. and Others}
}

@article{sofdetr2022,
title = {SOF-DETR: Improving small objects detection using transformer},
journal = {Journal of Visual Communication and Image Representation},
note = {Article S1047320322001432},
year = {2022},
author = {Gao, P. and Others}
}

@article{fasterilod2020,
title = {Faster ILOD: Incremental learning for object detectors based on Faster RCNN},
journal = {Pattern Recognition Letters},
note = {Article S0167865520303627},
year = {2020},
author = {Peng, C. and Others}
}

@article{applefaster2024,
title = {Detection model based on improved Faster-RCNN in apple orchard environment},
journal = {Smart Agricultural Technology},
year = {2024},
author = {Li, S. and Others}
}

@article{archreview2025,
title = {Architecture review: Two-stage and one-stage object detection},
journal = {Results in Engineering},
note = {Article S2773186325001100},
year = {2025},
author = {Kumar, A. and Others}
}

@article{domaininc2025,
title = {Domain incremental learning for object detection},
journal = {Pattern Recognition},
volume = {162},
pages = {111324},
year = {2025},
author = {Liu, B. and Others}
}

@article{selftarget2024,
title = {Learning self-target knowledge for few-shot segmentation},
journal = {Pattern Recognition},
volume = {149},
pages = {110236},
year = {2024},
author = {Li, Q. and Others}
}

@article{attnfpn2022,
title = {Attentional Feature Pyramid Network for small object detection},
journal = {Neural Networks},
note = {Article S089360802200329X},
year = {2022},
author = {Wang, J. and Others}
}

@article{smallobj2020,
title = {Recent advances in small object detection based on deep learning},
journal = {Image and Vision Computing},
note = {Article S0262885620300421},
year = {2020},
author = {Tong, K. and Others}
}

@misc{M-RCNN,
      title={Mask R-CNN}, 
      author={Kaiming He and Georgia Gkioxari and Piotr Dollár and Ross Girshick},
      year={2018},
      eprint={1703.06870},
      archivePrefix={arXiv},
      primaryClass={cs.CV},
      url={https://arxiv.org/abs/1703.06870}, 
}

@article{Revolutionizing2024,
  author = {Jafar, A. and Bibi, N. and Naqvi, R. A. and Sadeghi-Niaraki, A. and Jeong, D.},
  title = {Revolutionizing agriculture with artificial intelligence: plant disease detection methods, applications, and their limitations},
  journal = {Frontiers in Plant Science},
  volume = {15},
  pages = {1356260},
  year = {2024}
}

@article{Survey2020,
  author = {Lameski, P. and Zdravevski, E. and Trajkovik, V. and Kulakov, A.},
  title = {A survey of public datasets for computer vision tasks in precision agriculture},
  journal = {Computers and Electronics in Agriculture},
  volume = {178},
  pages = {105760},
  year = {2020}
}

@article{Song2021,
  author = {Song, P. and Wang, J. and Guo, X. and Yang, W. and Zhao, C.},
  title = {High-throughput phenotyping: Breaking through the bottleneck in future crop breeding},
  journal = {The Crop Journal},
  year = {2021},
  volume = {9},
  number = {3},
  pages = {633--645}
}

@article{Afonso2020,
  author = {Afonso, M. and Fonteijn, H. and Fiorentin, F. S. and others},
  title = {Tomato Fruit Detection and Counting in Greenhouses Using Deep Learning},
  journal = {Frontiers in Plant Science},
  year = {2020},
  volume = {11},
  pages = {571299}
}

@article{Wang2022,
  author = {Wang, C. and Liu, B. and Liu, L. and Zhu, Y. and Hou, J. and Liu, P. and Li, X.},
  title = {Application of Convolutional Neural Network-Based Detection Methods in Fresh Fruit Production: A Comprehensive Review},
  journal = {Frontiers in Plant Science},
  year = {2022},
  volume = {13},
  pages = {868745}
}

@article{Chapman2021,
  author = {Chapman, S.C. and others},
  title = {Scaling up high-throughput phenotyping for abiotic stress selection in the field},
  journal = {Theoretical and Applied Genetics},
  year = {2021},
  volume = {134},
  pages = {1845--1866}
}

@article{Zhu2024,
  author = {Zhu, H. and others},
  title = {Intelligent agriculture: deep learning in {UAV-based} remote sensing imagery for crop diseases and pests detection},
  journal = {Frontiers in Plant Science},
  year = {2024},
  volume = {15},
  pages = {1435016}
}

@article{Gill2022,
  author = {Gill, T. and Gill, S.K. and Saini, D.K. and Chopra, Y. and de Koff, J.P. and Sandhu, K.S.},
  title = {A comprehensive review of high throughput phenotyping and machine learning for plant stress phenotyping},
  journal = {Phenomics},
  year = {2022},
  volume = {2},
  number = {3},
  pages = {156--183}
}

@article{Visakh2024,
  author = {Visakh, V. R. and others},
  title = {Precision Phenotyping in Crop Science: From Plant Traits to Gene Discovery for Climate-Smart Agriculture},
  journal = {Plant Breeding},
  year = {2024},
  pages = {pbr.13228}
}

@article{PerezPatricio2024,
  author = {P{\'e}rez-Patricio, M. and others},
  title = {A systematic review of multi-mode analytics for enhanced plant stress evaluation},
  journal = {Frontiers in Plant Science},
  volume = {16},
  pages = {1545025},
  year = {2024}
}

@inproceedings{SAM,
  author = {Kirillov, Alexander and Mintun, Eric and Ravi, Nikhila and Mao, Hanzi and Rolland, Chloe and Gustafson, Laura and Xiao, Tete and Whitehead, Spencer and Berg, Alexander C. and Lo, Wan-Yen and Doll{\'a}r, Piotr and Girshick, Ross},
  title = {Segment Anything},
  booktitle = {Proceedings of the IEEE/CVF International Conference on Computer Vision (ICCV)},
  year = {2023}
}

@misc{RCNN,
  title = {Rich feature hierarchies for accurate object detection and semantic segmentation},
  author = {Ross Girshick and Jeff Donahue and Trevor Darrell and Jitendra Malik},
  year = {2014},
  eprint = {1311.2524},
  archivePrefix = {arXiv},
  primaryClass = {cs.CV}
}

@misc{DETR,
  title = {End-to-End Object Detection with Transformers},
  author = {Nicolas Carion and Francisco Massa and Gabriel Synnaeve and Nicolas Usunier and Alexander Kirillov and Sergey Zagoruyko},
  year = {2020},
  eprint = {2005.12872},
  archivePrefix = {arXiv},
  primaryClass = {cs.CV}
}

@misc{yolov8,
  title = {Real-Time Flying Object Detection with {YOLOv8}},
  author = {Dillon Reis and Jordan Kupec and Jacqueline Hong and Ahmad Daoudi},
  year = {2024},
  eprint = {2305.09972},
  archivePrefix = {arXiv},
  primaryClass = {cs.CV}
}

@misc{yolov9,
  title = {{YOLOv9}: Learning What You Want to Learn Using Programmable Gradient Information},
  author = {Chien-Yao Wang and I-Hau Yeh and Hong-Yuan Mark Liao},
  year = {2024},
  eprint = {2402.13616},
  archivePrefix = {arXiv},
  primaryClass = {cs.CV}
}

@misc{RNCDL,
  title = {Learning to Discover and Detect Objects},
  author = {Vladimir Fomenko and Ismail Elezi and Deva Ramanan and Laura Leal-Taix{\'e} and Aljo{\v{s}}a O{\v{s}}ep},
  year = {2022},
  eprint = {2210.10774},
  archivePrefix = {arXiv},
  primaryClass = {cs.CV}
}

@misc{DetPro,
  title = {Learning to Prompt for Open-Vocabulary Object Detection with Vision-Language Model},
  author = {Yu Du and Fangyun Wei and Zihe Zhang and Miaojing Shi and Yue Gao and Guoqi Li},
  year = {2022},
  eprint = {2203.14940},
  archivePrefix = {arXiv},
  primaryClass = {cs.CV}
}

@inbook{OV_DETR,
  title = {Open-Vocabulary {DETR} with Conditional Matching},
  isbn = {9783031200779},
  issn = {1611-3349},
  doi = {10.1007/978-3-031-20077-9\_7},
  booktitle = {Computer Vision -- ECCV 2022},
  publisher = {Springer Nature Switzerland},
  author = {Zang, Yuhang and Li, Wei and Zhou, Kaiyang and Huang, Chen and Loy, Chen Change},
  year = {2022},
  pages = {106--122}
}

@misc{GDINO,
  title = {Grounding {DINO}: Marrying {DINO} with Grounded Pre-Training for Open-Set Object Detection},
  author = {Shilong Liu and Zhaoyang Zeng and Tianhe Ren and Feng Li and Hao Zhang and Jie Yang and Chunyuan Li and Jianwei Yang and Hang Su and Jun Zhu and Lei Zhang},
  year = {2023},
  eprint = {2303.05499},
  archivePrefix = {arXiv},
  primaryClass = {cs.CV}
}

@misc{CFM,
  title = {Enhancing Novel Object Detection via Cooperative Foundational Models},
  author = {Rohit Bharadwaj and Muzammal Naseer and Salman Khan and Fahad Shahbaz Khan},
  year = {2023},
  eprint = {2311.12068},
  archivePrefix = {arXiv},
  primaryClass = {cs.CV}
}

@misc{dataset1,
  title = {Fruits Detection},
  howpublished = {\url{https://www.kaggle.com/datasets/andrewmvd/fruit-detection}},
  year = {2020}
}

@misc{dataset2,
  title = {Fruit Images for Object Detection},
  howpublished = {\url{https://www.kaggle.com/datasets/mbkinaci/fruit-images-for-object-detection}},
  year = {2018}
}

@misc{dataset3,
  title = {Fruit Object Detection},
  howpublished = {\url{https://www.kaggle.com/datasets/eunpyohong/fruit-object-detection}},
  year = {2021}
}

@misc{dataset4,
  title = {Fruits Images Dataset: Object Detection},
  howpublished = {\url{https://www.kaggle.com/datasets/afsananadia/fruits-images-dataset-object-detection}},
  year = {2024}
}

@misc{CLIP,
  title = {Learning Transferable Visual Models From Natural Language Supervision},
  author = {Alec Radford and Jong Wook Kim and Chris Hallacy and Aditya Ramesh and Gabriel Goh and Sandhini Agarwal and Girish Sastry and Amanda Askell and Pamela Mishkin and Jack Clark and Gretchen Krueger and Ilya Sutskever},
  year = {2021},
  eprint = {2103.00020},
  archivePrefix = {arXiv},
  primaryClass = {cs.CV}
}

@article{CropDeep2019,
  author = {Zheng, Yang-Yang and Kong, Jianlei and Jin, Xinbing and Wang, Xinyu and Zuo, Min},
  title = {{CropDeep}: The Crop Vision Dataset for Deep-Learning-Based Classification and Detection in Precision Agriculture},
  journal = {Sensors},
  volume = {19},
  number = {5},
  pages = {1058},
  year = {2019},
  doi = {10.3390/s19051058}
}

@misc{nawaz2024agriclip,
   title   = {{AgriCLIP}: Adapting {CLIP} for Agriculture and Livestock
              via Domain-Specialized Cross-Model Alignment},
   author  = {Umair Nawaz and Muhammad Awais and Hanan Gani and
              Muzammal Naseer and Fahad Khan and Salman Khan and
              Rao Muhammad Anwer},
   year    = {2024},
   eprint  = {2410.01407},
   archivePrefix = {arXiv},
   primaryClass  = {cs.CV}
 }

@inproceedings{stevens2024bioclip,
   title     = {{BioCLIP}: A Vision Foundation Model for the Tree of Life},
   author    = {Samuel Stevens and Jiaman Wu and Matthew J Thompson and
                Elizabeth G Campolongo and Chan Hee Song and
                David Edward Carlyn and Li Dong and Wasila M Dahdul and
                Charles Stewart and Tanya Berger-Wolf and
                Wei-Lun Chao and Yu Su},
   booktitle = {Proceedings of the IEEE/CVF Conference on Computer
                Vision and Pattern Recognition (CVPR)},
   pages     = {19412--19424},
   year      = {2024}
 }

@misc{gu2025bioclip2,
      title={BioCLIP 2: Emergent Properties from Scaling Hierarchical Contrastive Learning}, 
      author={Jianyang Gu and et al},
      year={2025},
      eprint={2505.23883},
      archivePrefix={arXiv},
      primaryClass={cs.CV},
      url={https://arxiv.org/abs/2505.23883}, 
}

@inproceedings{yang2024biotrove,
   title     = {{BioTrove}: A Large Curated Image Dataset Enabling
                {AI} for Biodiversity},
   author    = {Chih-Hsuan Yang and Benjamin Feuer and Zaki Jubery and
                Zi K Deng and Andre Nakkab and Md Zahid Hasan and
                Shivani Chiranjeevi and Kelly Marshall and
                Nirmal Baishnab and Asheesh K Singh and Arti Singh and
                Soumik Sarkar and Nirav Merchant and Chinmay Hegde and
                Baskar Ganapathysubramanian},
   booktitle = {Advances in Neural Information Processing Systems
                (NeurIPS)},
   volume    = {37},
   pages     = {102101--102120},
   year      = {2024}
 }

@misc{liu2024remoteclip,
      title={RemoteCLIP: A Vision Language Foundation Model for Remote Sensing}, 
      author={Fan Liu and Delong Chen and Zhangqingyun Guan and Xiaocong Zhou and Jiale Zhu and Qiaolin Ye and Liyong Fu and Jun Zhou},
      year={2024},
      eprint={2306.11029},
      archivePrefix={arXiv},
      primaryClass={cs.CV},
      url={https://arxiv.org/abs/2306.11029}, 
}

@misc{tschannen2025siglip2,
   title   = {{SigLIP 2}: Multilingual Vision-Language Encoders with
              Improved Semantic Understanding, Localization, and
              Dense Features},
   author  = {Michael Tschannen and Alexey Gritsenko and Xiao Wang and
              Muhammad Ferjad Naeem and Ibrahim Alabdulmohsin and
              Nikhil Parthasarathy and Talfan Evans and Lucas Beyer and
              Ye Xia and Basil Mustafa and Olivier H{\'e}naff and
              Jeremiah Harmsen and Andreas Steiner and Xiaohua Zhai},
   year    = {2025},
   eprint  = {2502.14786},
   archivePrefix = {arXiv},
   primaryClass  = {cs.CV}
 }

@misc{zhai2023sigmoid,
      title={Sigmoid Loss for Language Image Pre-Training}, 
      author={Xiaohua Zhai and Basil Mustafa and Alexander Kolesnikov and Lucas Beyer},
      year={2023},
      eprint={2303.15343},
      archivePrefix={arXiv},
      primaryClass={cs.CV},
      url={https://arxiv.org/abs/2303.15343}, 
}

@misc{inthewilds2023,
      title={PlantSeg: A Large-Scale In-the-wild Dataset for Plant Disease Segmentation}, 
      author={Tianqi Wei and Zhi Chen and Xin Yu and Scott Chapman and Paul Melloy and Zi Huang},
      year={2024},
      eprint={2409.04038},
      archivePrefix={arXiv},
      primaryClass={cs.CV},
      url={https://arxiv.org/abs/2409.04038}, 
}

@misc{benchmarkingwild2024,
      title={Benchmarking In-the-wild Multimodal Disease Recognition and A Versatile Baseline}, 
      author={Tianqi Wei and Zhi Chen and Zi Huang and Xin Yu},
      year={2024},
      eprint={2408.03120},
      archivePrefix={arXiv},
      primaryClass={cs.CV},
      url={https://arxiv.org/abs/2408.03120}, 
}

@misc{snapdiagnose2024,
      title={Snap and Diagnose: An Advanced Multimodal Retrieval System for Identifying Plant Diseases in the Wild}, 
      author={Tianqi Wei and Zhi Chen and Xin Yu},
      year={2024},
      eprint={2408.14723},
      archivePrefix={arXiv},
      primaryClass={cs.CV},
      url={https://arxiv.org/abs/2408.14723}, 
}

@misc{u3m2025,
      title={U3M: Unbiased Multiscale Modal Fusion Model for Multimodal Semantic Segmentation}, 
      author={Bingyu Li and Da Zhang and Zhiyuan Zhao and Junyu Gao and Xuelong Li},
      year={2024},
      eprint={2405.15365},
      archivePrefix={arXiv},
      primaryClass={cs.CV},
      url={https://arxiv.org/abs/2405.15365}, 
}

@misc{agrobench2025,
      title={AgroBench: Vision-Language Model Benchmark in Agriculture}, 
      author={Risa Shinoda and Nakamasa Inoue and Hirokatsu Kataoka and Masaki Onishi and Yoshitaka Ushiku},
      year={2025},
      eprint={2507.20519},
      archivePrefix={arXiv},
      primaryClass={cs.CV},
      url={https://arxiv.org/abs/2507.20519}, 
}

@misc{wacvagri2025,
      title={Leveraging Vision Language Models for Specialized Agricultural Tasks}, 
      author={Muhammad Arbab Arshad and Talukder Zaki Jubery and Tirtho Roy and Rim Nassiri and Asheesh K. Singh and Arti Singh and Chinmay Hegde and Baskar Ganapathysubramanian and Aditya Balu and Adarsh Krishnamurthy and Soumik Sarkar},
      year={2025},
      eprint={2407.19617},
      archivePrefix={arXiv},
      primaryClass={cs.LG},
      url={https://arxiv.org/abs/2407.19617}, 
}

\end{document}